\documentclass[a4paper, 10pt, conference]{IEEEtran}      
\IEEEoverridecommandlockouts

\usepackage{graphics} 
\usepackage{graphicx}
\usepackage{amsmath} 

\usepackage{amssymb}  
\usepackage{diagbox}
\usepackage{hyperref}
\usepackage{xcolor}

\title{\LARGE \bf Online Human Activity Recognition Employing Hierarchical Hidden Markov Models 
}

\author{Parviz Asghari$^{*}$, Elnaz Soleimani$^{*}$ and Ehsan Nazerfard$^{1}$

\thanks{$^{*}$P. Asghari and E. Soelimani are members of Ambient Intelligence Research Lab, Department of Computer Engineering and Information Technology,
 Amirkabir University of Technology, 424 Hafez Ave. Tehran, Iran.
        {\tt\small parviz.asghari@aut.ac.ir,}
        {\tt\small elnaz.soleimani@aut.ac.ir}}
\thanks{$^{1}$Department of Computer Engineering and Information Technology,
       Amirkabir University of Technology, 424 Hafez Ave. Tehran, Iran
        {\tt\small nazerfard@aut.ac.ir}}%
}

\providecommand{\keywords}[1]
{
  \textbf{\textit{Keywords---}} #1
}

\begin{document}

\maketitle
\thispagestyle{empty}
\pagestyle{empty}

\begin{abstract}

In the last few years there has been a growing interest in Human Activity Recognition~(HAR) topic. Sensor-based HAR approaches, in particular, has been gaining more popularity owing to their privacy preserving nature. Furthermore, due to the widespread accessibility of the internet, a broad range of streaming-based applications such as online HAR, has emerged over the past decades. However, proposing sufficiently robust online activity recognition approach in smart environment setting is still considered as a remarkable challenge. This paper presents a novel online application of Hierarchical Hidden Markov Model in order to detect the current activity on the live streaming of sensor events. Our method consists of two phases. In the first phase, data stream is segmented based on the beginning and ending of the activity patterns. Also, on-going activity is reported with every receiving observation. This phase is implemented using Hierarchical Hidden Markov models. The second phase is devoted to the correction of the provided label for the segmented data stream based on statistical features. The proposed model can also discover the activities that happen during another activity - so-called interrupted activities. After detecting the activity pane, the predicted label will be corrected utilizing statistical features such as time of day at which the activity happened and the duration of the activity. We validated our proposed method by testing it against two different smart home datasets and demonstrated its effectiveness, which is competing with the state-of-the-art methods.

\end{abstract}

\section{INTRODUCTION}
Nowadays, the prominent enhancement of sensor technology, alongside with the remarkable progress on Machine Learning methods has opened up a wide variety of real-world applications through combining different types of sensor and machine learning techniques for data acquisition  and processing, respectively. Various sensors have been utilized in research projects. Numerous studies have been done using visual sensors such as cameras which receive the sequence of images or videos as the sensor data. Despite the successful performance of vision based solutions,  non-visual sensors are still required to solve the existing limitations. Non-visual sensors can be installed on the human's body or in the environment as wearable and ambient sensors. Exploiting the network of heterogeneous sensors has been of interest in many recent studies with the aim of enhancing the performance as well.

Human Activity Recognition~(HAR) as an active research area in Machine Learning is of great importance. In many of the real world applications, system's decision is made based on the current human activity which is inferred by its HAR module from sensors data.

Smart Environment is a remarkable real world application which aims to monitor the behavior of individuals in the environment. This environment is enriched with sensors, actuator and processing units which can also be embedded in the objects. Monitoring systems as a use case of Smart Environments, are deployed to predict, detect and neutralize the abnormal behaviors of individuals in environments such as metro stations.

Ambient Assisted Living~(AAL) approaches use Activity Recognition methods in order to monitor resident's behavior and support them so that improve the quality of their lives. HAR has also been exploited in game consoles, fitness and health monitoring on smartphones' applications. Many research laboratories have paid considerable attention to the Smart Environment related topics and consequently, several research projects such as CASAS~\cite{cook2009collecting}, MAVhome~\cite{cook2003mavhome}, PlaceLab~\cite{intille2005placelab}, CARE~\cite{krose2008care}, and AwareHome~\cite{kidd1999aware} have been deployed. Activity recognition in this context denotes recognizing the daily activities of a person who is living in the environment. Activities of Daily Living (ADLs) is a terminology defined in the healthcare domain and it refers to people's daily self-care activities \cite{katz1970progress}.

\subsection{Activity Modeling}
In general, there exist two main approaches to model the activity; data-driven and knowledge-driven approach. Data-driven models, analyze large sets of data for each activity by means of machine learning and data mining techniques. Another approach is to acquire sufficient prior knowledge of personal preferences and tendencies by applying knowledge engineering and management technology methods.
Activity recognition task contains different challenges such as labeling the activities, interrupted activities, sensors with non-valuable data and activity extraction. However, online recognition is noted as one of the most dominant challenges since the main requirement in many modern real-world applications is to recognize user's behavior based on the online stream of data.

In the present study, the stream of data is generated from a set of binary environmental sensors' output.
The majority of online activity recognition approaches are fed with the data which preprocessed using the sliding window technique. The main drawback of this approach is the optimization problem of window size parameter itself.

To address this issue, we proposed a method to detect the activity pane on data stream by discovering the sequential pattern of the sensors data in the beginning and termination of each activity. Detecting the activity boundaries, we can recognize the activity using a proper segment of streaming sensor data. Since the beginning of an activity is detected, the model keep reporting the identified activity till the end of it gets recognized by its sequential pattern. While the model is reporting the current activity, it is capable of recognizing other activities interrupting the ongoing one. Having activity's segment identified, statistical features of the segment including activity duration will be taken as the input to correct predicted label of the activity. Consequently, we get ride of the non-valuable sensor data by ignoring all the intervals but activity panes and tuning the window size parameter.

Our model consists of a set of Hidden Markov Models implemented in a hierarchical manner in order to recognize the beginning, estimate on-going activity, detect end and class of the activities.
Our evaluation results demonstrated the superiority of the proposed model in comparison with some existing ones and competitive performance compare to the state-of-the-art models.

The remainder of the paper is organized as follows.
Section II outlines the previous researches have been done in this domain. Section III describes the proposed method and implementation details. Evaluation and experimental results are presented in Section IV and Section V summarizes the results of this work and draws conclusions.

\section{Related Work}

Smart environments obtain knowledge from physical platform and its residents in order to modify setting based on residents interactions' pattern with environment. These systems optimize numerous goals such as monitoring environment, assisting residents and  control resources. The first idea that uses sensor technology for creating smart environment backs to 1990. Moser's neural network house modifies environment conditional statues based on user behavior pattern \cite{mozer1998neural}. The next decades witnessed a considerable progress in this field which obtained by exploiting HAR techniques.

Several publications have appeared documenting research works on Human Activity Recognition. Generally, sensor-based and vision-based approaches, have been documented in the literature as two main categories of HAR. This paper is concentrating on the sensor-based HAR. The common basis for the majority of attempts of this type is to process the sequence of sensors' events and detect the corresponding activity, based on the discovered pattern. Furthermore, activity recognition has been vastly used in various applications and environments. Hence, it is conspicuous to have a robust Activity Recognition Model that be invariant to the environmental settings such as sensors' structure and placement in the smart environment.

To address this requirement, author in~\cite{cook_learning_2012} proposed a robust model of the sensors which extracts different features such as the duration of the activity. In this generalized model, the problem is independent of the sensor environment and can be implemented for different users. Besides, to evaluate the models, having enough correctly labeled data is inevitable. Nevertheless, manual labeling of the sensors' data, is prohibitive due to being highly time-consuming and often inaccurate.

Moreover, the way that annotation has been done is often ignored while it makes bias on the data. In the most common approach of annotating, residents of the smart environment are asked to perform an activity and then annotation will be done based on activated sensors. Nonetheless, this approach may not be practical in all situations. A labeling mechanism presented in~\cite{szewcyzk2009annotating} is an example of existing solutions for annotating sensor data automatically.

Much researches on activity recognition have been done using pre-segmented data; which means the beginning and end of the activities is pre-determined in the dataset~\cite{sanchez2008activity,fleury2009supervised}.  Such approaches are far unrealistic compared to real-world setting and are not applicable in the online applications as the beginning and end of the activities is not determined when it comes to the stream of data. Researchers in~\cite{hong2013segmenting,okeyo2014dynamic} developed methods for sensor stream segmentation which brings the activity recognition based systems closer to those of actual world.

Despite all the researches on HAR, still opened challenges such as overlapping or concurrent activities, have yet to be solved. Overlap is noted as the phenomenon that different activity classes activate the same set of sensor events which makes overlapping activities  hard to discriminate only based on the types of sensor events that they have triggered~\cite{wen2015activity}. AALO is an Activity recognition system in presence of overlapped activities which works based on Active Learning. It can recognize the overlapped activities by preprocessing step and item-set mining phase~\cite{hoque2012aalo}. A key limitation of this research is that it is not capable of recognizing the overlapping activities which happen in the same location. In addition, performing on the online data stream is still lacking in this study.

Moreover, researchers in~\cite{malazi2018combining} suggest a two-phase method based on emerging pattern which can recognize complex activities.In the first phase, this method extracts emerging pattern for distinct  activities. In the second phase, it segments streaming sensor data, then uses time dependency between segments in order to concatenate the relevant segments. Segments concatenation lets the method recognize complex activities. In~\cite{quero2018real}, Quero et al. proposed a Real-Time method for recognizing interleaved activities based on Fuzzy Logic and Recurrent Neural Networks.

Authors in~\cite{cook2013activity,gjoreski2017unsupervised} studied the problem of handling the large proportion of available data that are not categorized in predefined classes and addressed it by discovering patterns in them and segmenting it into learnable classes. These kinds of data usually belong to the sensors that are not exclusively involved in the predefined class of activities.

Several Machine Learning approaches have been examined in the domain of Activity Recognition. Ensemble methods~\cite{jurek2014clustering}, non-parametric models~\cite{sun2014nonparametric}, Temporal Frequent pattern mining~\cite{nazerfard2018temporal}, SVM-based models~\cite{sanchez2008activity}, Recurrent Neural Networks~\cite{singh2017human}, and probabilistic models like Hidden Markov Model and the Markov Random Field~\cite{kabir2016two,kasteren2010activity,yan2016real} have been exploited in the literature. Nonetheless, less attention has been paid to the domain of Online Activity Recognition which deals with processing stream of sensor data contrary to the conventional approaches that utilize pre-segmented data.

Most of the presented solutions for streaming data processing are based on sliding window technique~\cite{okeyo2014dynamic,krishnan2014activity,yala2015feature,al2016windowing,al2018activity}.
The sliding window approach, briefly named as \textit{windowing}, mainly considers the temporal relation or number of sensors for framing data. One of the key bottlenecks of this approach is fine-tuning the window size. One basic solution is to employ constant pre-determined  window size~\cite{kasteren2010activity,krishnan2014activity}. Though, as the number of activated sensors are varied in different activities, applying dynamic window size have been noticed by many researchers~\cite{wan2015dynamic,espinilla2018new,al2016windowing,okeyo2014dynamic}.

Authors in~\cite{al2016windowing,al2018activity} present a novel probabilistic method to determine the window size. A different window size is initialized regarding each class of activity based on prior estimation and it is getting updated by the upcoming sensor events.

Krishnan et al.~\cite{krishnan2014activity} consider temporal dependency among sensors as a criterion for sliding windowing so that those sensors, which get activated at a certain time interval, are examined as one activity. Their other proposed solution is a sensor-based method which recognizes group of sensors as a window that were continuously activated together by their Mutual Information measure.
Authors in~\cite{yala2015feature} extended the presented model of~\cite{krishnan2014activity} and improved its  performance by altering computation of Mutual Information.

Researchers in~\cite{kabir2016two} presented a multi-stage classification method. The first stage is to cluster the activities using a Hidden Markov Model based on location data and then in the next stage, another HMM classifies the exact activity using a sequence of sensor data. The major weakness of this method is that it makes no attempt to specify the boundaries of activities which negatively affects the performance.

Another method to tackle online recognition is introduced by Li et al.~\cite{li2018time}. They proposed cumulative  fixed  sliding windows for real time activity recognition. Their segmentation method consists of several fixed time length windows which have overlapped with each other. These overlapping windows considered as the whole a window, and its information is used to detect the on-going activities.

\section{Proposed approach}

The Windowing technique is among the most widely used solutions to process the stream of sensor data. However, one practical question arises when dealing with this approach is how to determine the proper window size. There is still some controversy surrounding window size tuning while it has a great impact on the performance as the decision-making process is halting until the model receives a complete window of data.
  \begin{figure}[tp]
\setlength{\fboxrule}{0pt}
      \framebox{
      \parbox{0.5\textwidth}{
      \centering
      \includegraphics[width=0.5\textwidth, height=12cm, keepaspectratio]{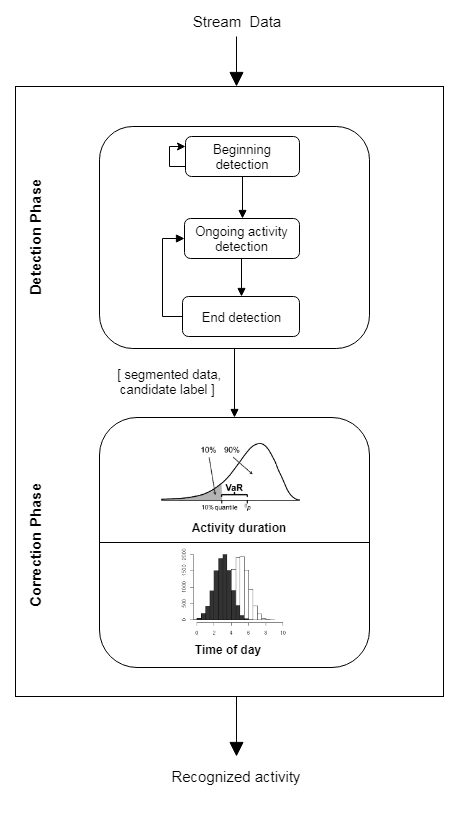}}}
      \caption{Abstract structure of the proposed method}
      \label{ost_structure}

   \end{figure}

\begin{figure}[bp]
\setlength{\fboxrule}{0pt}
\framebox{\parbox{0.49\textwidth}{
\centering
\includegraphics[width=0.49\textwidth,height=10cm, keepaspectratio]{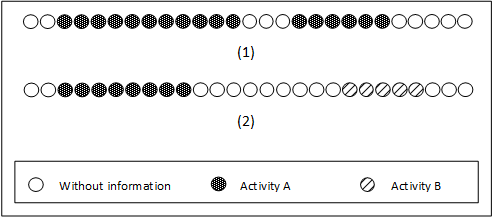}}}
  \caption{(1) distinct samples with different sizes (2) different classes have different sizes.}
  \label{window_size}
\end{figure}

Moreover, the selected window passed to the model often contains inappropriate data which belongs to none of the predefined classes. Dynamic sliding window size may seem an appropriate solution. However, as it is shown in Fig.~\ref{window_size}, not only different classes of activity need different window sizes, but also distinct samples of one class require distinct window sizes.

This paper seeks to address these problems by presenting a new model which works based on the dynamic segmentation of the data stream. Our proposed model breaks down the problem of Online Activity Recognition into two sub-problems as it is depicted in Fig.~\ref{ost_structure}. In the initial stage, the model detects the activity pane by recognizing sensor events' patterns of when the activities start and finish and makes the prediction while receiving data. Having the beginning and end of the activities determined, in the next stage the model proceeds to refine the prediction by classifying the activities into the predefined classes and the \textit{other} class. Predefined classes proposed by medical specialists ~\cite{katz1970progress}, included Daily Living Activities such as eating, bathing, sleeping, etc. \textit{Other} class label is devoted to the activities that do not fit into predefined classes. Following sections explain each stage in detail.

The main contributions and innovations of this paper are as follows:

\begin{description}
  \item[$\bullet$] Description of a novel method based on the occurrence pattern of activities, that addresses window size problem of previous sliding window methods.
  \item[$\bullet$] Providing an estimation of the on-going activity due to each sensor observation, which delivers Real-time Activity Recognition.
  \item[$\bullet$] Recognizing interrupted activities even those which occur in the same location. \\
\end{description}

\subsection{Detection Phase}
\begin{figure}[bp]
\setlength{\fboxrule}{0pt}
  \centering
  \framebox{\parbox{0.49\textwidth}{\includegraphics[width=0.49\textwidth]{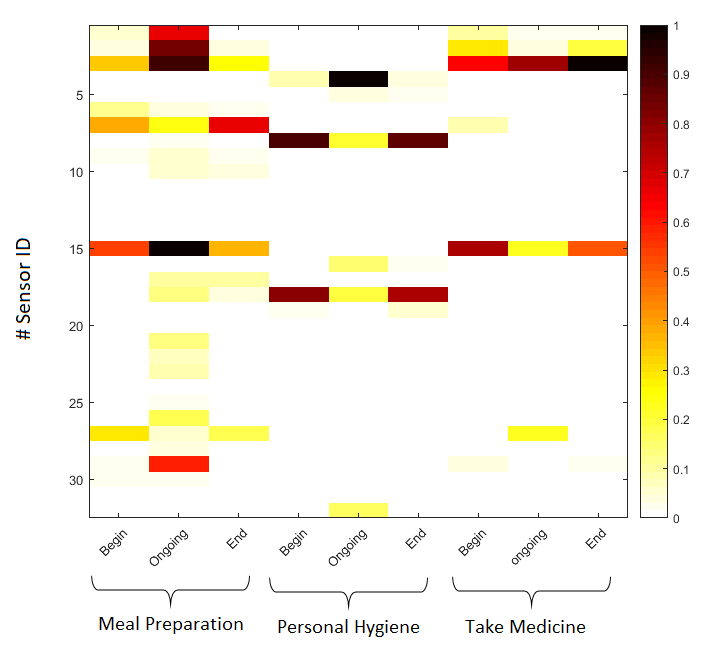}}}
  \caption{Normalized frequency of sensor activation in the beginning, end and duration of \textit{Meal Preparation}, \textit{Personal Hygiene} and \textit{Take Medicine} activities.}
  \label{activation_freq}
\end{figure}

The first step to recognize the activity on the stream of sensor data is to detect its occurrence. The idea is to investigate the sensor events in the beginning and end of the activities to extract corresponding patterns of start and end. As it can be inferred from Fig.~\ref{activation_freq}, the set of activated sensors, for each activity is almost unchanged in most of the occurrences. This assumption can be justified by considering locally limited functional zone for each activity. \textit{Meal preparation} as an example happens in the kitchen, thus it should be kitchen's sensors that get activated during this activity.

The Number of observations which are taken into account for beginning detection purpose, affects the model performance. Table \ref{table_sensor_size} compares different values for this parameter and corresponding results. Based on this comparison, we opted for considering 3 sensor events in our model. Indeed, the number of considered sensors should be small enough to promise the recognition feasibility of the next activity. To exemplify, when the resident opens or closes the door, only a few sensors get activated; hence waiting for more observations is out of options.

\begin{table}[t]
\caption{Proposed model accuracy\% obtained using \#$\beta$ observations}
\label{table_sensor_size}
\begin{center}
\begin{tabular}{l|llll}
\diagbox{Dataset}{$\beta$} & 2 & 3 & 4 & 5 \\
\hline
\hline
\\[-1.5ex]
HomeA  & 78.7  & \textbf{97}  & 92.9  & 93.2  \\
HomeB & 84.3  & \textbf{96.4}  & 94.2  &91
\end{tabular}
\end{center}
\end{table}

\begin{figure}[t]
\setlength{\fboxrule}{0pt}
  \centering
  \framebox{\parbox{0.48\textwidth}{\includegraphics[width=0.48\textwidth]{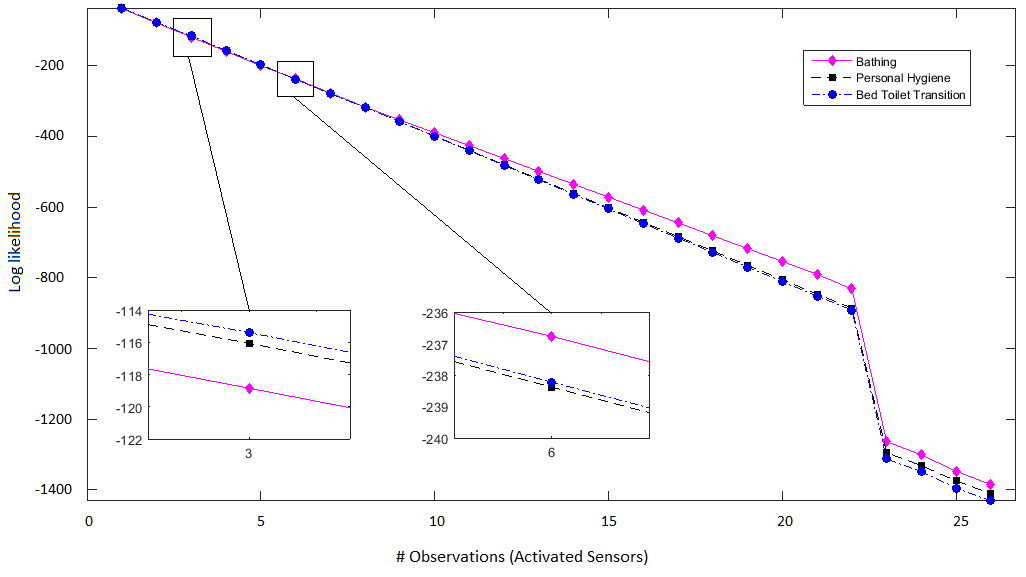}}}
  \caption{Log likelihood trend of candidate classes during the observation procedure}
  \label{likelihood_trend}
\end{figure}

On the other hand, there exist some classes of activities such as \textit{Personal Hygiene} and \textit{Bathing}, which share a common set of activated sensors in the beginning and the end. Therefore, recognizing these activities requires more information like sensor activation sequences. Once the beginning of an activity gets recognized, the model utilizes the upcoming sequence of sensor events to recognize the activity class itself and as it receives more events, its prediction gradually becomes more confident. This trend can be seen in Fig.~\ref{likelihood_trend}.

As Fig.~\ref{likelihood_trend} shows, at first the most probable activity is \textit{Bed to Toilet transition}, while after receiving more observations, estimated activity changes to \textit{Bathing}. These activities have overlap due to their similar sensors. Therefore, there is a need for more information to produce a reliable prediction.

Numerous Machine Learning methods such as SVM, Neural Networks, Decision Tree, and Probabilistic Graphical Models  have been applied in activity recognition field. Hidden Markov Model is one of the most popular models in the literature for sequential data processing as it does not suffer from labor-intensive calculations like Neural Networks. Besides, prior knowledge can be simply leveraged in probabilistic models which are also more robust to the noise.

Taking into account the aforementioned reasons, we chose HMM to recognize the activities. Suppose $X_{t}$ represents the hidden state vector and $Y_{t}$ represents vector of observations. Assuming K  possible hidden states we have $X_{t}\in \{1,...,K\}$. In our case, observations are the sequence of sensor events. Equation \eqref{eq:eq_hmm} highlights the calculation to obtain the most probable class based on the observations:

\begin{equation}
\label{eq:eq_hmm}
P(X_{t}|y_{1:t}) = P(y_{t}|X_{t}) [\sum_{X_{t-1}} P(X_{t}|x_{t-1}) P(x_{t-1}|y_{1:t-1}) ]
\end{equation}

The Hidden Markov Model approach is not well suited to process long sequences and this led us to employ  an extension of HMM  called Hierarchical Hidden Markov Model in our study. Hierarchical-HMM is suitable for the problems that contain multilevel dependencies in terms of time and follow a hierarchical structure in their context.

Hierarchical-HMM is represented by a triple $(\zeta,\gamma, \theta)$ while $\zeta$, $\gamma$, and $\theta$ stand for model structure, set of observations and model's parameters respectively. Model structure defines number of levels $d$ and children-parent relations on each level. States in the lowest level, are the only generative ones which generate the observations. Parameter set $\theta$ is defined as following~\cite{nguyen2005learning}:

\begin{equation}
\label{eq:eq_theta}
\theta = \{B(y|p), \pi^{d,p^*},  A_{i, j}^{d,p^*}, A_{i, end}^{d,p^*}   |   \forall(y,p,d,p^*,i,j) \},
\end{equation}

where $p^*$ is an abstract state in level $d$ with children set $ch(p^*)$, $B(y|p)$ is the probability of generating observation $y$ by generative state $p$, $\pi^{d,p^*}$ is the prior probability of $ch(p*)$, $A_{i, j}^{d,p^*}$ is the probability of inhering child $i$ to $j$, and $A_{i, end}^{d,p^*}$ is the probability of termination of $p^*$ in state $i$. The proper value set of $\theta$ is obtained by Maximum Likelihood technique.

\begin{figure*}[t]
    \centering

  \framebox{\parbox{0.98\textwidth}{
    \centering
  \includegraphics[width=0.98\textwidth,height=5cm, keepaspectratio]{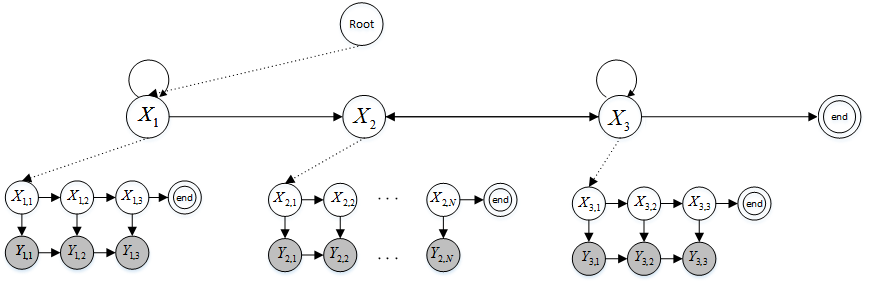}}}
  \centering
  \caption{The transition diagram of the implemented Hierarchical-HMM}
  \label{hhmm}
\end{figure*}

Fig.~\ref{hhmm} demonstrates the transition diagram of our implemented Hierarchical-HMM. In this diagram we have 2 type of transition , vertical and horizontal. Vertical transitions are shown by dotted arc, and occurs between different level of HHMM. Horizontal transitions happen among same level node, when task of one node finishes.

Additionally, there are two types of node in HHMM, abstract nodes which produce a sequence of observations, and production nodes which produce single observation. In Fig.~\ref{hhmm}, nodes $X_{1}$, $X_{2}$, and $X_{3}$ are abstract nodes that each of them considered as a distinct HMM. In this model, $X_{1}$ is the responsible sub-HMM for detecting the beginning of an activity. $X_{1}$ is an auto regressive  HMM which consist of 3 observation. After receiving 3 sensor observations, if beginning of an activity is detected, the control will pass through the $X_{2}$ in order to recognize the class of on-going activity.

In $X_{2}$ sub-HMM with occurring each observation, the corresponding activity will be recognized. This process continues until the termination of activity which detects by $X_{3}$. At the end of the cycle, control will be given back to the root node. The output of this process $S$, is the segmentation of data which holds the related information.

\begin{figure}[b]

  \centering
  \framebox{\parbox{0.47\textwidth}{\includegraphics[width=0.47\textwidth]{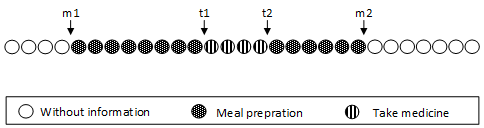}}}
  \caption{An illustration on interrupted activities: The subject has been preparing meal since time \textit{m1}. At time \textit{t1} he decided to interrupt so as to take his medicine. At time \textit{t2} he resumed and kept preparing meal till \textit{m2}.}
  \label{interrupted_sample}
\end{figure}

Sometimes, subject interrupts the current activity and commences a new activity. There exist some techniques which can recognize the latter activity independently of the first one~\cite{nef2015evaluation}. Thought, their discrimination is limited to the activities which happen at different locations. Since our hierarchical detection technique is location invariant the proposed model is capable of recognizing interrupted activities that take place in a common location. Fig.~\ref{interrupted_sample} demonstrates an example of the interrupted activities.

\subsection{Correction Phase}

\begin{figure}[t]
\setlength{\fboxrule}{0pt}
\framebox{\parbox{0.47\textwidth}
{
\centering
\includegraphics[width=0.47\textwidth,height=10cm, keepaspectratio]{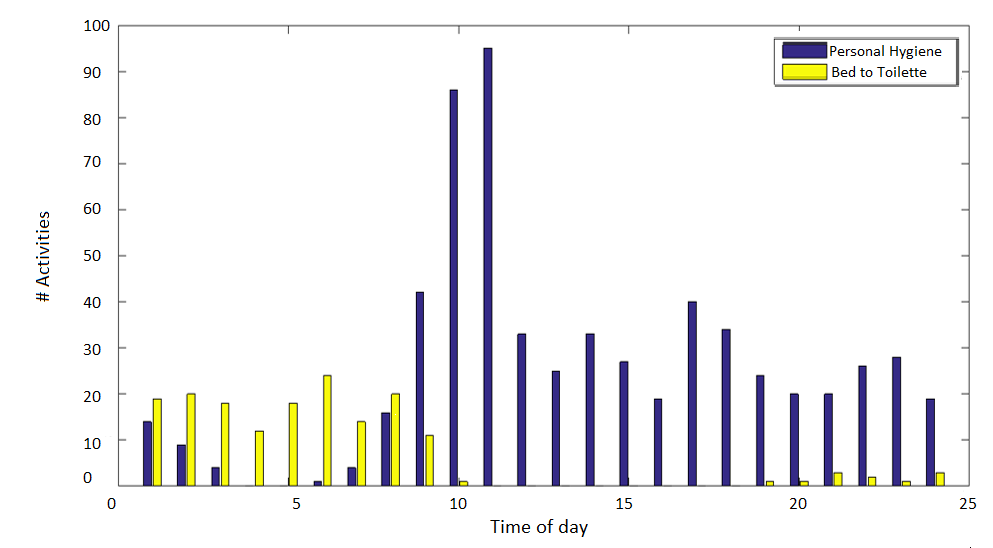}}}
  \caption{Choosing time of day as a feature facilitates classification task. \textit{Bed to Toilette transition} and \textit{Personal Hygiene} activities can be discriminated by this feature since \textit{Bed to Toilette transition} occurrence time has low overlap with the \textit{Personal Hygiene} activity.}
  \label{feature_selection}
\end{figure}

With the completion of the Detection phase, our model can move to the next stage. In this stage, for each class, Joint Probabilistic Distribution Function (Joint PDF) of its time length and time of day gets extracted. Fig.~\ref{feature_selection} exemplifies how time of the day as an statistical feature can ease the discrimination. The extracted PDFs now can be used to measure the level of belonging of a data segment to each of the classes. As Equation~\eqref{eq:eq_prob} reveals, if this probability passes the threshold $\alpha$ for each class $C_{k}$, the data can be labeled as $C_{k}$. Otherwise, it is labeled as Other class. A fine tuning process has been done to achieve the proper value for $\alpha$. Table~\ref{table_alpha_tuning} details this trial.

\begin{equation}
\label{eq:eq_prob}
P( S \in C_k )  =
\left\{
\begin{array}{ll}
    1   &   f(T_{C_{k}}=t_{S}) \geq \alpha\\
    0   &  \text{Otherwise}
\end{array} \right.
\end{equation}

\begin{table}[t]
\caption{Accuracy of the proposed model employing different threshold parameter $\alpha$}
\label{table_alpha_tuning}
\begin{center}
\begin{tabular}{l|lllll}
\diagbox{Dataset}{$\alpha$ } & 0.02 & 0.04 & 0.06 & 0.08 & 0.10 \\
\hline
\hline
\\[-1.5ex]
HomeA  & 59.7  & 62.1  & 63  & \textbf{65.2} &64.8 \\
HomeB & 54  & 53  & \textbf{60} &58 & 57.4
\end{tabular}
\end{center}
\end{table}

\begin{table}[t]
\caption{Dataset Statistics}
\label{dataset_characterictics}
\begin{center}
\begin{tabular*}{0.9\columnwidth}{@{\extracolsep{\fill}}lllll}

\textbf{\# Attribute}      & \textbf{HomeA}            & \textbf{HomeB}             \\
\hline
\hline \\[-1.5ex]
Motion sensors~         & \multicolumn{1}{r}{20}     & \multicolumn{1}{r}{18}      \\
Door sensors            & \multicolumn{1}{r}{12}     & \multicolumn{1}{r}{12}      \\
Residents               & 1 Person                  & 1 Person                    \\
Sensor events collected & \multicolumn{1}{r}{371925} & \multicolumn{1}{r}{274920}  \\
\\[-1.5ex] \hline \\[-1.5ex]
Timespan                & 5 months                    & 5 months

\end{tabular*}
\end{center}
\end{table}

In summary, this model first detects the activity occurrence utilizing the beginning pattern of activities. Next, it recognizes the class of ongoing activity based on activated sensors until the end of the activity gets detected. To improve the discrimination performance between similar activities, our model also exploits the time of day in which the activity is occurring as an statistical feature.

\section{Evaluation}

The performance of our proposed approach is assessed by conducting 2 experiments on 2 different datasets to recognize the current activity on the stream of data. We compared our approach with a set of most common Machine Learning models and achieved convincing results.

\subsection{Dataset}

\begin{figure}[bp]
\setlength{\fboxrule}{0pt}
  \centering
  \framebox{\parbox{0.48\textwidth}{\includegraphics[width=0.48\textwidth]{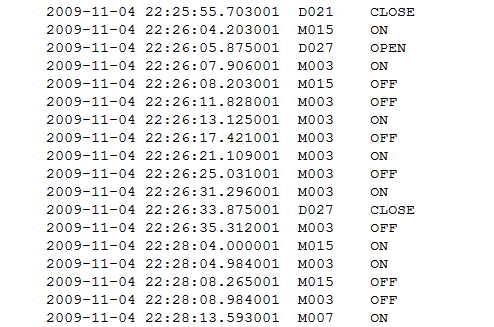}}}
  \caption{Sample of sensor data during an
  activity in dataset HomeA and HomeB}  \label{smaple_of_dataset}
\end{figure}

\begin{figure}[tp]
\setlength{\fboxrule}{0pt}
\framebox{\parbox{0.49\textwidth}{
\centering
\includegraphics[width=0.49\textwidth,height=10cm, keepaspectratio]{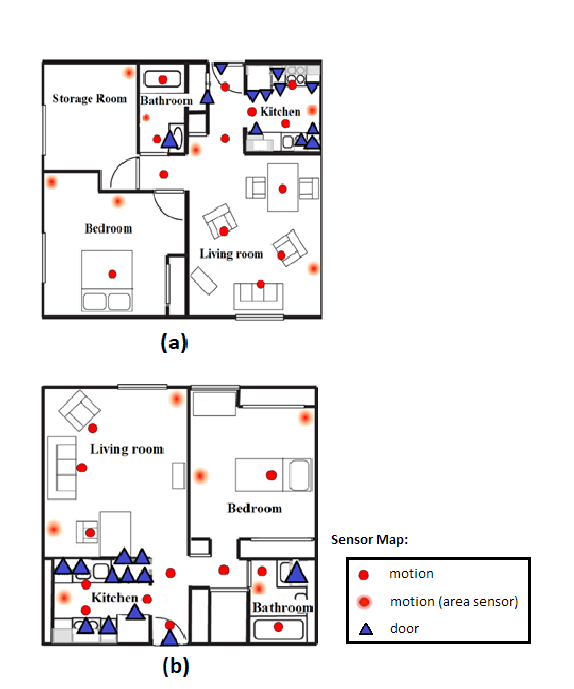}}}
  \caption{The architecture of (a) HomeA (b) HomeB.  Red circles represent  motion sensors and blue triangles stand for door sensors.}
  \label{home_arch}
\end{figure}

We evaluated our proposed model on HomeA\footnote{\url{http://eecs.wsu.edu/~nazerfard/AIR/datasets/data1.zip}} and HomeB\footnote{\url{http://eecs.wsu.edu/~nazerfard/AIR/datasets/data2.zip}} dataset. Fig.~\ref{smaple_of_dataset} shows a sample slice of this dataset. These 2 datasets have been collected using 32 and 30 sensors which have been deployed in 2 different houses depicted in Fig. \ref{home_arch} for the period of 5 months. Labeling of this dataset has been done later by human experts. Table~\ref{dataset_characterictics} summarizes characteristics of data acquisition for HomeA and HomeB datasets.
These datasets contain 11 classes of interest. Details on sample distribution of classes are highlighted in Table \ref{dataset_classes}.

\begin{table*}[t]
\caption{Confusion matrix of the proposed model on HomeA dataset}
\label{table_f1_home1}
\begin{center}
\begin{tabular*}{\textwidth}{l|@{\extracolsep{\fill}}llllllllllll}
~                           & 1    & 2    & 3    & 4    & 5    & 6    & 7    & 8    & 9    & 10   & 11   & 12    \\
\hline \hline \\[-1.5ex]
1=Personal Hygiene          & 0.88 & 0.00 & 0.00 & 0.02 & 0.00 & 0.00 & 0.00 & 0.02 & 0.00 & 0.00 & 0.00 & 0.08  \\
2=Leave Home                & 0.00 & 0.14 & 0.00 & 0.00 & 0.29 & 0.00 & 0.00 & 0.00 & 0.00 & 0.00 & 0.00 & 0.57  \\
3=Enter Home                & 0.00 & 0.30 & 0.14 & 0.00 & 0.00 & 0.00 & 0.03 & 0.00 & 0.00 & 0.00 & 0.00 & 0.54  \\
4=Bathing                   & 0.08 & 0.00 & 0.00 & 0.67 & 0.00 & 0.00 & 0.00 & 0.00 & 0.00 & 0.00 & 0.00 & 0.25  \\
5=Meal Preparation          & 0.00 & 0.00 & 0.00 & 0.00 & 0.52 & 0.00 & 0.08 & 0.00 & 0.04 & 0.00 & 0.00 & 0.36  \\
6=Napping                   & 0.00 & 0.00 & 0.02 & 0.00 & 0.01 & 0.13 & 0.00 & 0.12 & 0.00 & 0.00 & 0.00 & 0.72  \\
7=Take Medicine             & 0.00 & 0.00 & 0.00 & 0.00 & 0.13 & 0.00 & 0.55 & 0.00 & 0.05 & 0.00 & 0.00 & 0.27  \\
8=Eating Food               & 0.00 & 0.00 & 0.01 & 0.00 & 0.01 & 0.00 & 0.00 & 0.28 & 0.00 & 0.00 & 0.00 & 0.69  \\
9=Housekeeping             & 0.00 & 0.00 & 0.00 & 0.00 & 0.13 & 0.00 & 0.26 & 0.00 & 0.00 & 0.00 & 0.00 & 0.61  \\
10=Sleeping in Bed          & 0.00 & 0.00 & 0.00 & 0.00 & 0.00 & 0.00 & 0.00 & 0.00 & 0.00 & 0.40 & 0.02 & 0.58  \\
11=Bed to Toilet transition & 0.60 & 0.00 & 0.00 & 0.00 & 0.00 & 0.00 & 0.00 & 0.00 & 0.00 & 0.00 & 0.27 & 0.13  \\
12=Other                    & 0.19 & 0.01 & 0.25 & 0.02 & 0.04 & 0.01 & 0.13 & 0.03 & 0.01 & 0.01 & 0.02 & 0.28
\end{tabular*}
\end{center}
\end{table*}

\begin{table*}[ht]
\caption{F1-measure of the proposed model on HomeB dataset}
\label{table_f1_home2}
\begin{center}
\begin{tabular*}{\textwidth}{l|@{\extracolsep{\fill}}llllllllllll}
~                           & 1    & 2    & 3    & 4    & 5    & 6    & 7    & 8    & 9    & 10   & 11   & 12    \\
\hline  \hline \\[-1.5ex]
1=Personal Hygiene          & 0.57 & 0.00 & 0.00 & 0.04 & 0.01 & 0.00 & 0.00 & 0.00 & 0.00 & 0.00 & 0.08 & 0.31  \\
2=Leave Home                & 0.00 & 0.50 & 0.00 & 0.00 & 0.0  & 0.00 & 0.00 & 0.00 & 0.00 & 0.00 & 0.00 & 0.50  \\
3=Enter Home                & 0.00 & 0.65 & 0.09 & 0.00 & 0.03 & 0.00 & 0.00 & 0.00 & 0.00 & 0.00 & 0.00 & 0.21  \\
4=Bathing                   & 0.00 & 0.00 & 0.00 & 1.0  & 0.00 & 0.00 & 0.00 & 0.00 & 0.00 & 0.00 & 0.00 & 0.00  \\
5=Meal Preparation          & 0.00 & 0.01 & 0.00 & 0.00 & 0.30 & 0.00 & 0.02 & 0.00 & 0.08 & 0.00 & 0.00 & 0.59  \\
6=Napping                   & 0.00 & 0.00 & 0.00 & 0.00 & 0.00 & 0.13 & 0.00 & 0.05 & 0.00 & 0.00 & 0.00 & 0.82  \\
7=Take Medicine             & 0.00 & 0.00 & 0.00 & 0.00 & 0.11 & 0.00 & 0.09 & 0.00 & 0.56 & 0.00 & 0.00 & 0.23  \\
8=Eating Food               & 0.00 & 0.00 & 0.00 & 0.00 & 0.00 & 0.00 & 0.00 & 0.33 & 0.04 & 0.00 & 0.00 & 0.63  \\
9=Housekeeping             & 0.00 & 0.00 & 0.00 & 0.00 & 0.05 & 0.00 & 0.00 & 0.00 & 0.35 & 0.00 & 0.00 & 0.60  \\
10=Sleeping in Bed          & 0.00 & 0.00 & 0.00 & 0.00 & 0.00 & 0.00 & 0.00 & 0.00 & 0.00 & 0.96 & 0.00 & 0.04  \\
11=Bed to Toilet transition & 0.13 & 0.00 & 0.00 & 0.03 & 0.00 & 0.00 & 0.00 & 0.00 & 0.00 & 0.00 & 0.56 & 0.28  \\
12=Other                    & 0.14 & 0.01 & 0.05 & 0.01 & 0.00 & 0.00 & 0.00 & 0.00 & 0.01 & 0.01 & 0.05 & 0.71
\end{tabular*}
\end{center}
\end{table*}

\begin{table}[!htbp]
\caption{Overall instances of each class in dataset HomeA and HomeB}
\label{dataset_classes}
\begin{center}
\begin{tabular}{lrr}
\\[-1.5ex]
\textbf{Activity}        & \multicolumn{1}{l}{\textbf{HomeA}} & \multicolumn{1}{l}{\textbf{HomeB}}  \\
\hline \hline \\[-1.5ex]
Personal Hygiene         & 704                                 & 782                                  \\
Leave Home               & 420                                 & 245                                  \\
Enter Home               & 417                                 & 245                                  \\
Bathing                  & 72                                  & 41                                   \\
Meal Preparation         & 554                                 & 245                                  \\
Napping                  & 564                                 & 138                                  \\
Take Medicine            & 473                                 & 81                                   \\
Eating Meal              & 334                                 & 208                                  \\
Housekeeping             & 13                                  & 175                                  \\
Sleeping in Bed          & 210                                 & 207                                  \\
Bed to Toilet transition & 96                                  & 236                                  \\
\\[-1.5ex] \hline \\[-1.5ex]
TOTAL                    & 2603                                & 3857
\end{tabular}
\end{center}
\end{table}

\begin{table}[t]
\caption{Performance comparison of models in terms of accuracy and F1-measure}
\label{table_results}
\begin{center}
\centering
\begin{tabular*}{\columnwidth}{l|@{\extracolsep{\fill}}llll}
       & \multicolumn{2}{c}{Accuracy}                          & \multicolumn{2}{c}{F1-measure}                         \\ \\[-1.5ex] \hline \\[-1.5ex]
Model  & \multicolumn{1}{l}{HomeA} & \multicolumn{1}{l}{HomeB} & \multicolumn{1}{l}{HomeA} & \multicolumn{1}{l}{HomeB}  \\
\\[-1.5ex] \hline \hline \\[-1.5ex]
Proposed model     & 65.20\%   &60\%  &51\%  &49\%  \\
SW        & 58\%   & 48\%      &51\%  &49\% \\
TW        & 59\%   & 52\%      &53\%  &54\% \\
SWMI      & 64\%   & 54\%      &60\%  &57\% \\
SWTW      & 62\%   & 45\%      &55\%  &55\% \\
DW        & 59\%   & 55\%      &58\%  &58\% \\
PWPA      & 62\%   & 50\%      &55\%  &51\%
\end{tabular*}

\end{center}
\end{table}

\subsection{Baseline models}
We have compared performance of our proposed model with several existing approaches:
\begin{itemize}
  \item SW: This model utilises constant window size. Sensors are equally contributing in window size calculation.

  \item TW: It employs a constant window size that is obtained based on time interval.

  \item SWMI: It considers window size as a constant value. Mutual Information of sensors and activity classes is taken into account for window size calculation.

  \item SWTW: Window size remains constant in this model and is calculated by time based weighting of the sensor events.

  \item DW: In this model window size is variable for each activity and its value is calculated using probabilistic information of sensor events.

  \item PWPA: It benefits from a 2-level window with a fixed sizes. First level contains the probabilistic information of  the previous windows and activities. Second level includes the data of current step with possible activity as a augmented feature.

\end{itemize}

\subsection{Experimental Results and Analysis}
We compare the performance of the models in terms of F1-measure and Accuracy as it is reported in Table \ref{table_f1_home1} ,\ref{table_f1_home2} and  \ref{table_results} respectively.
For HomeA dataset, \textit{Personal Hygiene}, \textit{Bathing}, \textit{Sleeping}, \textit{Take Medicine}, \textit{Meal Preparation}, and \textit{Eating Meal} classes can be highly discriminated. However, unsatisfactory results were reached for \textit{Enter Home}, \textit{Leave Home}, and \textit{Sleeping} classes. This performance degradation is justifying given the fact that \textit{Enter Home} and \textit{Leave Home} both activate the same set of sensors and these two classes of activity often occur in sequence. Besides, sometimes the resident walks out of the house just for short moments which is not considered as Leaving home.

In HomeA dataset, \textit{Bathing} and \textit{Sleeping} are well discriminated and detection of \textit{Meal Preparation}, \textit{Personal Hygiene}, \textit{Eating}, and \textit{Bed to Toilette transition} are satisfactory. The performance dropped for \textit{Taking Medicine}, \textit{Napping}, \textit{Enter Home} and \textit{Leave Home}.
According to Table \ref{dataset_classes}, \textit{Housekeeping} activity has only 13 instances in the HomeA dataset which is dramatically low compared to the rest of the classes. This imbalanced distribution may have been the cause of inadequate performance of the model.

Generally, classes of activity that have the same local domain such as \textit{Napping} and \textit{Eating Meal}, are tough to discriminate. Note that the long time gap can compensate local overlap as it does in distinguishing \textit{Personal Hygiene} and \textit{Bed to Toilette transition}.

\section{Conclusions and Future Work}
Robust online activity recognition in the domain of smart environments is considered as one of the most dominant challenges. While most of the previous approaches suffer from window size parameter tuning, this paper has highlighted a new activity recognition approach on sensor data stream which determines the beginning and end of the activities. In addition, our proposed model is capable of recognizing interrupted activities. The experimental results of this study indicate the efficiency of the proposed model compared to the existing ones.

In general, the accuracy of the activity recognition methods cannot exceed a certain amount due to the uncertainty of human behavior. Nevertheless, uncertain predictions can be offered through a recommendation context in smart environments.

Further work needs to be performed to achieve improvement on recognition of \textit{Other} class as the most problematic activity class. Current recognition systems require a fair amount of labeled data to reach a certain satisfactory accuracy while data acquisition and labeling in this domain is impractical. Future studies on the current topic are therefore  recommended in the domain of Knowledge Transfer to reduce the crucial need of labeled data.

\addtolength{\textheight}{-11cm}


\bibliographystyle{ieeetr}
\bibliography{references}

\end{document}